\documentclass[english]{article}
\usepackage{graphicx}
\usepackage{amsmath}
\usepackage[table]{xcolor}
\usepackage{subcaption}
\usepackage{float}
\usepackage[margin=2.5cm]{geometry}
\usepackage[hyphens]{url}
\usepackage{hyperref}
\usepackage{amsthm,amsfonts,amssymb,amsmath}
\usepackage{graphicx}
\usepackage{xcolor}
\usepackage{bbm}
\usepackage{paralist}
\usepackage{enumitem}

\usepackage[labelfont=bf,labelsep=period]{caption}

\usepackage[nameinlink,capitalise,noabbrev]{cleveref}
\crefformat{equation}{#2(#1)#3}

\providecommand{\keywords}[1]
{
  \small	
  \textbf{\textit{Keywords---}} #1
}

\let\originalleft\left
\let\originalright\right
\renewcommand{\left}{\mathopen{}\mathclose\bgroup\originalleft}
\renewcommand{\right}{\aftergroup\egroup\originalright}

\begin{document}

\title{Multi Kernel Estimation based Object Segmentation}
\author{Haim Goldfisher \and Asaf Yekutiel}
\date{22 October 2024\\[1ex]\footnotesize Project Repository: \url{https://github.com/kuty007/Multi-Kernel-GAN}}

\maketitle
\begin{abstract}
This paper presents a novel approach for multi-kernel estimation by enhancing the KernelGAN algorithm \cite{KernelGAN}, which traditionally estimates a single kernel for the entire image. We introduce Multi-KernelGAN, which extends KernelGAN's capabilities by estimating two distinct kernels based on object segmentation masks. Our approach is validated through three distinct methods: texture-based patch Fast Fourier Transform (FFT) calculation, detail-based segmentation, and deep learning-based object segmentation using YOLOv8 \cite{YOLO} and the Segment Anything Model (SAM) \cite{SAM}. Among these methods, the combination of YOLO and SAM yields the best results for kernel estimation. Experimental results demonstrate that our multi-kernel estimation technique outperforms conventional single-kernel methods in super-resolution tasks.
\end{abstract} \hspace{10pt}


\section{Introduction}
\subsection{Super-Resolution Problem Overview}
Super-Resolution (SR) represents a crucial challenge in image processing that has garnered significant attention in both research and practical applications, including image enhancement. The objective of the SR process is to improve the quality of images by reconstructing detailed high-resolution patterns that closely resemble their original forms from a degraded low-resolution (LR) image \( I_{LR} \) to yield a high-resolution (HR) image \( I_{HR} \). Typically, in the context of image SR studies, \( I_{LR} \) is subjected to degradation due to a blur kernel \( k_{blur} \) and the presence of additive noise. This relationship can be mathematically described as illustrated in \autoref{eq:lr}.

\begin{equation}
\label{eq:lr}
I_{LR} = (I_{HR} * k_{blur}) \downarrow_{s} + n
\end{equation}

\subsection{Blind and Non-Blind SR Methods}
Super-resolution (SR) methods are broadly categorized into non-blind and blind approaches, each with its distinct methodology.

Non-blind SR methods assume that the degradation model, such as the blur kernel or noise, is known or fixed. These methods rely on large amounts of training data, typically learning the super-resolution process directly from high and low-resolution image pairs. Often based on deep learning, non-blind SR models map low-resolution images to their high-resolution counterparts. While these methods can produce impressive results, their performance tends to decline when real-world degradations differ from those seen during training.

Blind SR methods, by contrast, focus on more challenging scenarios where the degradation model is unknown. Instead of relying on pre-trained models or fixed degradation assumptions, blind SR methods aim to estimate the degradation process directly from the test image. These methods often exploit statistical patterns within the image itself. The work by Irani, Shocher, and Cohen demonstrated that nearly every image contains repetitive patterns (e.g., 5$\times$5 or 7$\times$7 patches) across different scales or regions. These patterns enable super-resolution from a single image without the need for external datasets, relying on internal image statistics, a process known as Single Image Super-Resolution (SISR) \cite{SISR}.

\subsection{ZSSR and KernelGAN}
ZSSR ("Zero-Shot" Super-Resolution) \cite{ZSSR} incorporates this approach using an internal convolutional neural network (CNN) that learns to map low-resolution (LR) images to their high-resolution (HR) counterparts. The method works by downscaling the LR image using bicubic interpolation and then training a CNN to reconstruct the LR image. This trained network is then applied to generate the HR image. Notably, ZSSR has outperformed data-driven non-blind methods, such as VDSR and EDSR+ \cite{VDSR,EDSR}. However, the limitation of ZSSR is that it uses bicubic interpolation as the default downscaling kernel, which may not match the actual degradation process.

KernelGAN \cite{KernelGAN} addresses this issue by learning the degradation kernel from the internal distribution of the test image itself, utilizing the architecture of Generative Adversarial Networks (GANs) \cite{GANs} to achieve this. This estimated kernel is then fed into ZSSR to replace the bicubic kernel, enhancing the super-resolution (SR) process and improving the final resolution output.

While blind SR methods are more adaptable, most rely on estimating a single kernel for the entire image. This can be insufficient for complex images with diverse textures and objects, where different regions may exhibit distinct degradation patterns. Addressing this limitation calls for more advanced techniques that go beyond single-kernel approaches, potentially enabling multi-kernel models that adapt to varying characteristics within an image for more robust and accurate SR results

\subsection{Motivation}
Our work introduces Multi-KernelGAN, an extension of KernelGAN that leverages object segmentation masks to estimate multiple kernels. By dividing an image into different segments and applying separate kernels, we improve the accuracy and robustness of kernel estimation.

While KernelGAN operates under the assumption that \textbf{different downscaling kernels are needed for different images}, our approach assumes that \textbf{different downscaling kernels are required for different regions within the same image}. This allows us to capture the varying characteristics of different regions, leading to more precise super-resolution results.

However, assigning a separate kernel for each pixel would not be effective, as it would make the process overly sensitive to noise and rapid changes in the image. Such an approach would lack robustness and could lead to inconsistencies in kernel estimation. Instead, by using segmentation masks, we strike a balance between flexibility and stability, ensuring that each region receives an appropriate kernel without overfitting to local variations.

\section{Methodology}
\subsection{Overview of Multi-KernelGAN}
Multi-KernelGAN extends the original KernelGAN by introducing binary masks for segmenting an image into two regions. Each region is assigned its own kernel, enabling the model to handle regions with different textures or structures.

\begin{figure}[htbp]
    \centering
    \includegraphics[width=.75\textwidth]{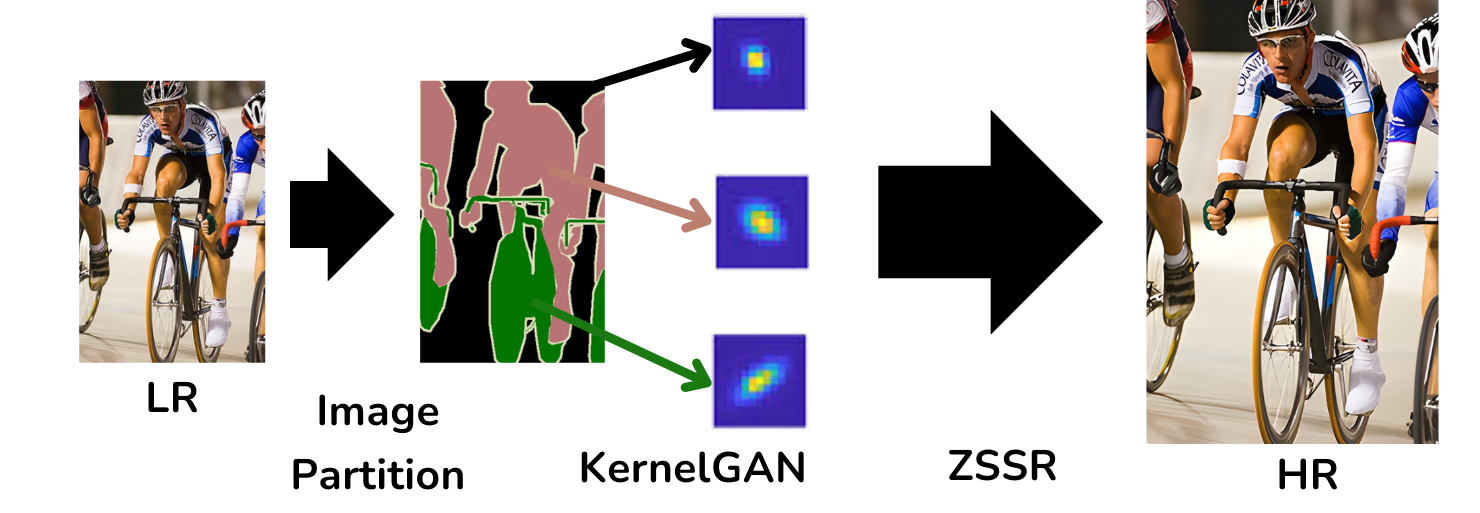}
    \caption{Multi-KernelGAN Model Pipeline.}
    \label{fig:modelPipeline}
\end{figure}

The model pipeline in \autoref{fig:modelPipeline} consists of the following stages:
\begin{enumerate}[label=\Roman*.]
    \item \textbf{Image Segmentation}: The input image is segmented into multiple regions using a mask.
    \item \textbf{Kernel Estimation}: Each region is associated with its own kernel, estimated through the GAN framework.
    \item \textbf{Super-Resolution}: Using the estimated kernels, ZSSR is applied separately to each region to achieve super-resolution.
\end{enumerate}

\subsection{Region-Based Mask Generation}

The region-based mask generation in Multi-KernelGAN is essential for dividing the input image into distinct regions that require different handling for texture and structure. By assigning each region its own kernel, the model can better capture variations in image details, which is key for achieving high-quality super-resolution.

The process begins with \textbf{object detection}, where we use a pre-trained YOLOv8 \cite{YOLO} model to identify key objects or areas in the image. This object detection helps differentiate between various regions, such as foreground and background, that are likely to have different textures or levels of detail.

\begin{figure}[htbp]
    \centering
    \includegraphics[width=.8\textwidth]{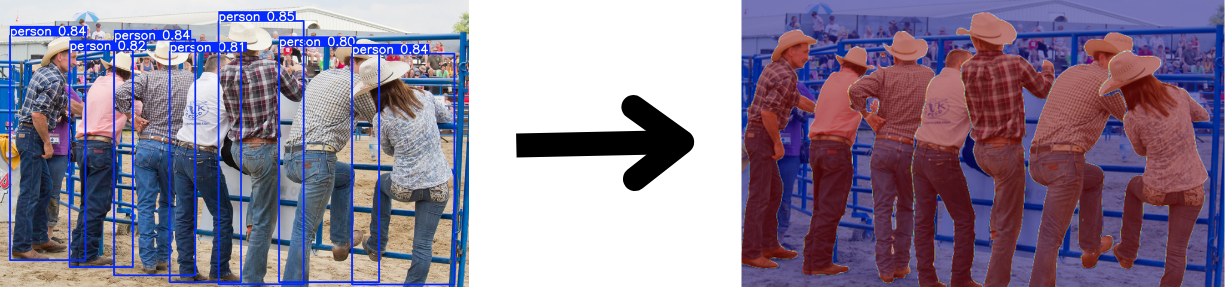}
    \caption{YOLOv8 and SAM-based Mask Creation. The YOLOv8 bounding boxes provide object localization, while SAM refines these regions with accurate segmentation.}
    \label{fig:YOLO+SAM}
\end{figure}

Once objects are detected, \textbf{segmentation masks} are generated for each detected region using the SAM \cite{SAM} predictor. These masks highlight the boundaries of the objects within the image, and the regions that fall outside these masks are treated as background. To ensure precise masking, SAM provides pixel-level accuracy, segmenting objects from the surrounding area. This process is illustrated in \autoref{fig:YOLO+SAM}, where YOLO's bounding boxes lead to accurate segmentation by SAM.

After generating the masks, a \textbf{binary mask} is constructed for the entire image. This binary mask acts as a layer that splits the image into foreground (regions with objects) and background. By separating these areas, the model can handle each region with its own unique processing strategy, which is crucial for handling diverse textures like sharp edges and smooth backgrounds.

Next, a \textbf{downscaling process} is applied to the binary mask. This involves resizing the mask to a lower resolution using nearest-neighbor interpolation, which introduces a "blocky" effect. The blocky mask simplifies the super-resolution task by reducing the complexity of the regions while retaining the overall structure. This scaled-down mask is then upscaled back to the original resolution, preserving its blocky nature for the subsequent stages.

Finally, the resulting masks for both the \textbf{foreground (object regions)} and \textbf{background} are saved separately. These masks guide the kernel estimation phase, where each region gets its own kernel, allowing for tailored super-resolution based on the region’s characteristics.

\subsection{Multi-KernelGAN}
\subsubsection{ Learning the Ideal Kernel for Each Region}
Once the regions are segmented, \textbf{KernelGAN} is applied to each region. KernelGAN is responsible for learning the ideal \textbf{downsampling kernel} that best describes the degradation of each region. This kernel is critical in determining how to best enhance the resolution of that particular region.

\begin{itemize}
    \item \textbf{KernelGAN} works in a self-supervised manner to learn the kernel directly from the low-resolution region patches, without requiring external training data. Each region is treated as a separate problem, allowing the network to learn an individualized kernel that optimally reconstructs the region.
\end{itemize}

\subsubsection{ZSSR for Region-Based Super-Resolution}
After KernelGAN learns the ideal kernels for each region, super-resolution is performed using \textbf{ZSSR}. ZSSR further refines the region by exploiting internal image information and applying the learned kernel to upscale each segment.

\begin{itemize}
    \item The super-resolution process is executed separately for each region, preserving the characteristics of each part of the image. ZSSR operates on patches from the same region, using internal self-similarity to reconstruct finer details.
\end{itemize}

\subsubsection{Reconstructing the Whole Image}
After applying KernelGAN + ZSSR to each region, the super-resolved regions are combined to form the final \textbf{super-resolved (SR) image}. The process of reconstruction involves merging the SR regions back together, ensuring seamless integration of the upscaled regions. This step is crucial in producing a high-quality super-resolution image that retains the details and characteristics learned from each region.

\subsubsection{Advantages of Multi-KernelGAN}
The Multi-KernelGAN approach allows for finer control over image upscaling by treating different regions with unique kernels and upscaling strategies. This results in a more accurate and detailed super-resolution image, as it avoids applying a uniform kernel across the entire image, which may lead to suboptimal results in heterogeneous regions.

\section{Previous Attempts}
\subsection{Global Frequency-Domain Texture Segmentation Using FFT}
\begin{figure}[htbp]
    \centering
    \includegraphics[width=1\textwidth]{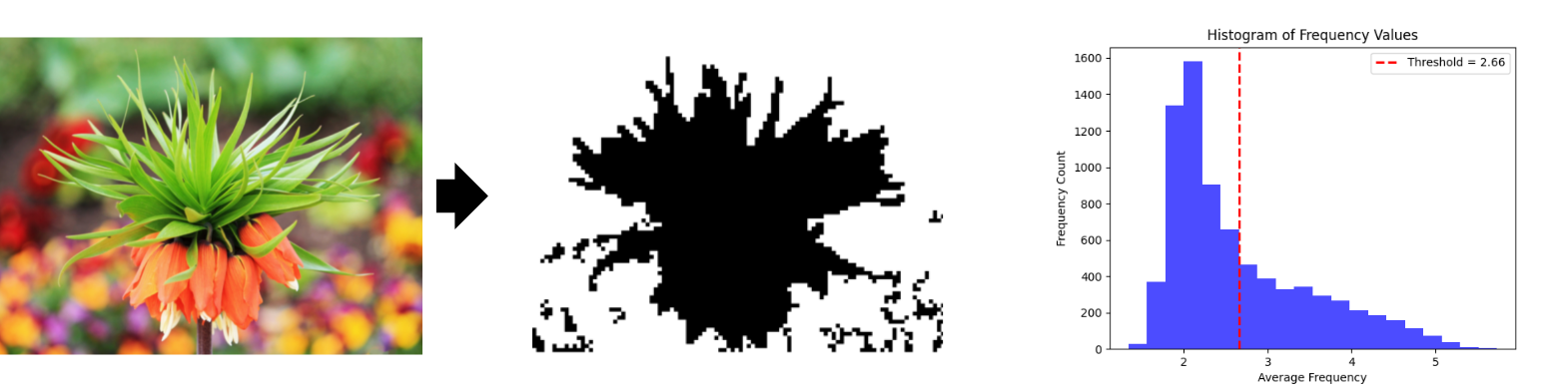}
    \caption{Frequency-Based Mask Using Fast Fourier Transform (FFT). The figure also shows the histogram of frequency values with a threshold: values to the left of the threshold map to Mask A (background), and values to the right map to Mask B (foreground).}
    \label{fig:fft_example}
\end{figure}

The patch-based Fast Fourier Transform (FFT) approach computes the frequency representation of image patches to segment objects from backgrounds based on their texture characteristics. By leveraging FFT, we aim to distinguish between high-frequency textures (often associated with objects) and low-frequency areas (commonly associated with smoother backgrounds), as can be seen in \autoref{fig:fft_example}. This approach capitalizes on the relationship between texture and object-background segmentation, where objects typically exhibit more complex textures than their surroundings.

\textbf{Key Steps:}
\begin{enumerate}[label=\Roman*.]
    \item \textbf{Frequency Representation}: For each image patch, the FFT is computed to transform the spatial domain into the frequency domain, providing insights into the texture patterns present in each patch.
    \item \textbf{Magnitude Spectrum Averaging}: The average magnitude spectrum of each transformed patch is calculated, which helps identify dominant frequency components.
    \item \textbf{Binary Mask Creation}: Based on the average frequency for each patch, a binary mask is constructed. Patches with frequencies above the overall average are marked as white (255), representing higher-frequency regions typically corresponding to object textures. Lower-frequency patches remain black (0), representing smoother background areas.
\end{enumerate}

While this approach performed well on simple images with homogeneous textures, it introduced noise when applied to more complex scenes. In such cases, isolated noise islands or artifacts appeared in the masks, which could be mitigated through post-processing techniques such as applying a blur filter or removing small, isolated regions based on neighborhood connectivity. \textbf{Despite these corrections, the resulting segmentation was not precise enough for more intricate object structures, limiting its overall accuracy.}

\subsection{Local Frequency-Based Texture Segmentation}
This segmentation approach utilizes two distinct methods to analyze texture details and identify areas of high information content within the image:

\renewcommand{\theenumi}{\thesubsection.\arabic{enumi}}
\begin{enumerate}[label=\theenumi]
    \item \textbf{Edge and Contour Detection:} This method highlights regions where significant texture transitions occur. By applying edge detection techniques and refining the resulting contours, we create a mask that emphasizes the detailed boundaries of objects in the image. This mask effectively segments the image into regions based on edge intensity and texture boundaries, expanding prominent areas while smoothing out smaller, irrelevant features.

    \item \textbf{Anchor Pixel Identification:} This method identifies anchor pixels by locating patches within the image that contain high concentrations of data-rich information. By computing the gradient magnitudes of small image patches, regions with significant variations in intensity are selected as anchor points. These areas, which exhibit strong local gradients, are crucial for representing the fine details of the image. The resulting mask from this approach highlights key areas of the image where detailed information is concentrated.
\end{enumerate}

\begin{figure}[htbp]
    \centering
    \includegraphics[width=1\textwidth]{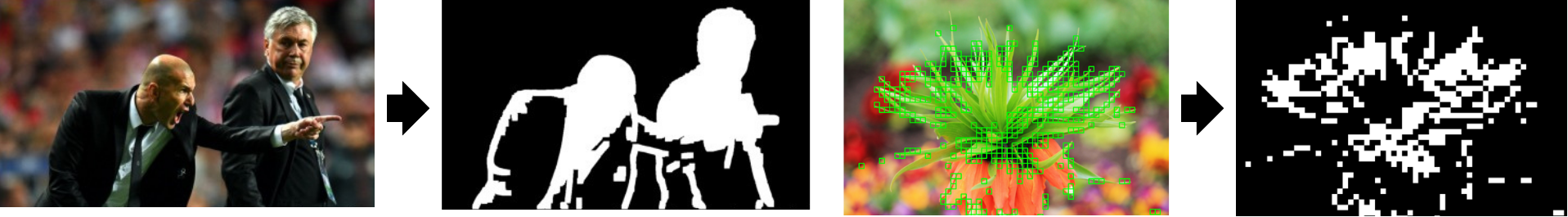}
    \caption{Details-Based Mask: From left to right, edges and contours mask, and creation of anchor pixels mask based on data-rich patches.}
    \label{fig:Texture-Details}
\end{figure}

Despite the advantages of these methods, they exhibited significant limitations in practical applications. The masks generated from both edge and contour detection, as well as anchor pixel identification, were often too small and noisy. This resulted in poor segmentation results, with many masked regions exhibiting strange colors along the edges. These artifacts arose due to the small size of the detected features, leading to a lack of clarity and definition in the masks.

Attempts to enlarge the masks to mitigate these issues only compounded the problem, as it caused most of the mask to encompass regions with non-rich or less detailed information. Consequently, while aiming for a clearer segmentation, the enlarged masks often included irrelevant areas, diminishing the overall effectiveness of the segmentation process. This highlights the need for more robust techniques that can accurately capture detailed textures without introducing unwanted noise or artifacts.

\subsection{Deep Learning-Based Object Segmentation}

Object segmentation using deep learning models has evolved significantly, and we tested several state-of-the-art methods to achieve accurate segmentation. Initially, we applied region-based CNN (R-CNN and Faster R-CNN) \cite{RCNN, FASTER_RCNN} and then moved to more advanced models like Detectron2 and SAM (Segment Anything Model) \cite{Detectron2, SAM}. However, each approach had notable limitations, which led us to the final strategy discussed in the previous section.

\begin{enumerate}
\renewcommand{\theenumi}{\thesubsection.\arabic{enumi}}
    \item \textbf{R-CNN and Faster R-CNN}: R-CNN \cite{RCNN} and Faster R-CNN \cite{FASTER_RCNN} provided object detection and segmentation by proposing regions of interest (ROIs), but they presented several issues:
    \begin{itemize}
        \item \textbf{Speed}: Both models, especially R-CNN, were computationally expensive and too slow for real-time or large-scale applications.
        \item \textbf{Precision}: The bounding boxes produced by these models were often too coarse, failing to capture finer object details and complex boundaries.
    \end{itemize}
    Given the need for precise, pixel-level segmentation, these models were inadequate for our use case.

    \item \textbf{Detectron2}: Detectron2 \cite{Detectron2} brought significant improvements over Faster R-CNN in terms of speed and accuracy, offering more detailed segmentation masks. However, we encountered challenges with:
    \begin{itemize}
        \item \textbf{Boundary Precision}: Detectron2 still struggled with precise object boundaries, especially when multiple objects overlapped or when objects were of similar size.
        \item \textbf{Complex Scenes}: In complex or cluttered environments, the masks were often incomplete or inaccurate.
    \end{itemize}
    Although Detectron2 offered better results than earlier models, its limitations in handling fine boundaries and complex scenes prompted us to explore other options.

    \item \textbf{SAM (Segment Anything Model)}: The Segment Anything Model (SAM) \cite{SAM} was tested for its ability to generate accurate masks at the pixel level. While SAM excels in segmenting complex objects without needing explicit object detection, it introduced new challenges. SAM generates multiple potential segments, making it difficult to determine the number of objects to segment without prior knowledge of the scene.
    
    As shown in \autoref{fig:SAM}, SAM tends to prioritize the largest segment, which in some cases, results in incomplete segmentation (e.g., only one object being captured). This ambiguity makes SAM less effective in cases where multiple objects of similar size are present.

    \begin{figure}[htbp]
        \centering
        \includegraphics[width=1\textwidth]{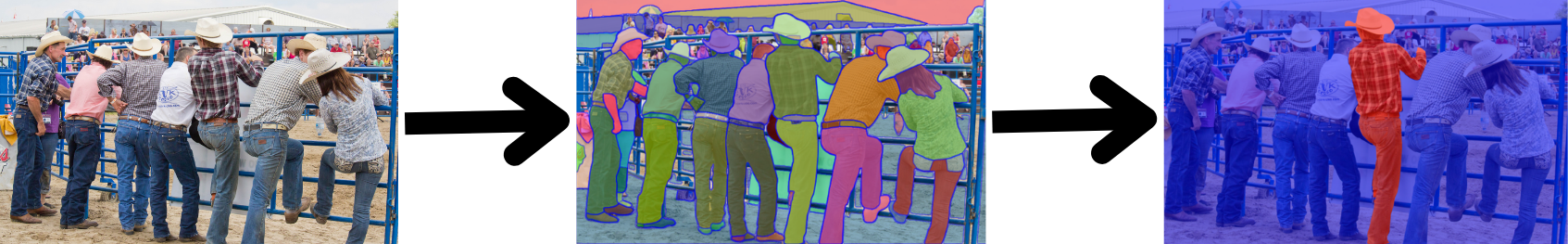}
        \caption{SAM Mask Creation. Selecting only the largest segment can result in incomplete segmentation, as seen in the image where one cowboy is captured, but others are missed.}
        \label{fig:SAM}
    \end{figure}

    Given SAM's difficulty in consistently capturing all objects without a clear boundary or object count, it was evident that SAM alone was insufficient for our segmentation task.
\end{enumerate}

To overcome the limitations of these methods, we adopted the \textbf{YOLO + SAM} approach, as detailed in the previous section. This combination effectively solves the object-count ambiguity and boundary precision issues by using YOLO \cite{YOLO} for object detection and SAM for detailed segmentation.

\section{Experiments and Results} 
\subsection{Dataset}
We conducted experiments using the DIV2K dataset \cite{DIV2K} that contain complex images with multiple textures and objects. The images were downsampled using different blur kernels to simulate real-world scenarios.

\subsection{Evaluation Metrics}
The evaluation metrics used in this experiment were Peak Signal-to-Noise Ratio (PSNR), Structural Similarity Index Measure (SSIM), and Mean Squared Error (MSE), and Visual Quality metrics. Additionally, qualitative comparisons were performed to assess how well the method captured different textures and structures.

\subsection{Results}

In this study, we evaluated the performance of two super-resolution techniques: \textit{Multi-KernelGAN+ZSSR} and \textit{KernelGAN+ZSSR}. Our dataset consisted of 50 images that conformed to our methodology, where each image contained a prominent central object that could be isolated from the background. This structure allowed us to apply segmentation techniques efficiently, facilitating the performance of super-resolution on both object and background components.

The results are summarized in Table \ref{tab:results}, presenting the average values for each metric across all sampled images.
\begin{figure}[htbp]
    \centering
    \includegraphics[width=.75\textwidth]{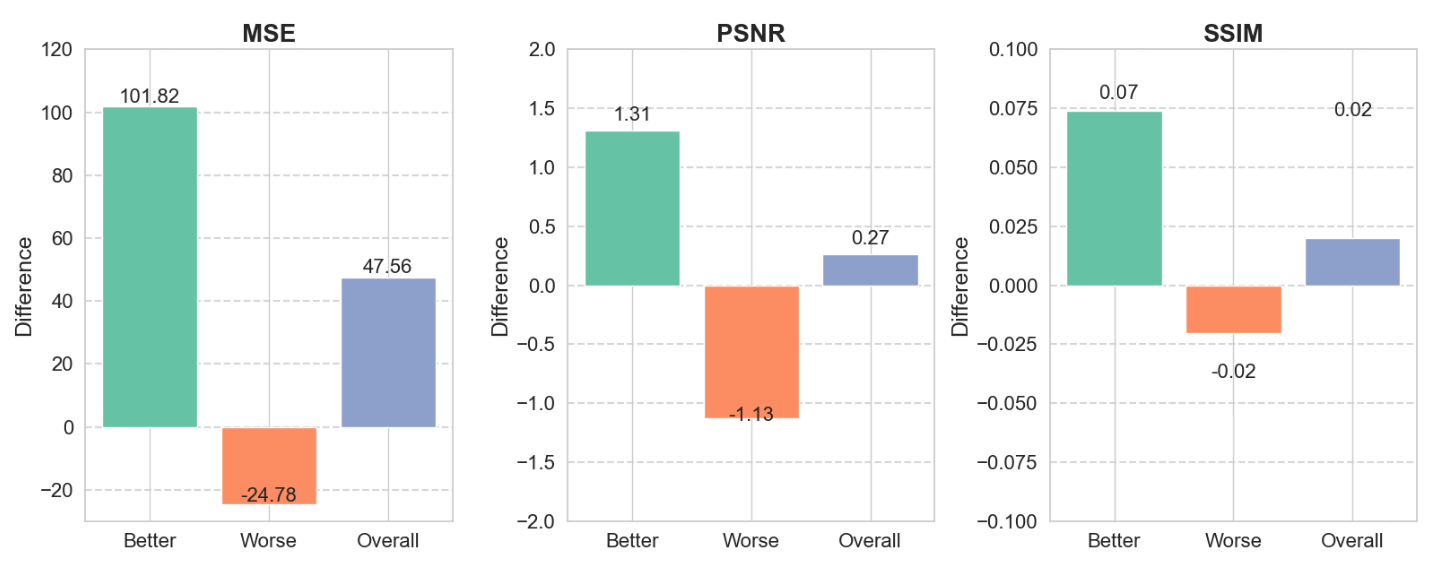}
    \caption{Average performance differences between the Multi KernelGAN images and the ZSSR+KernelGAN.
where positive values are for cases  where Multi-KernelGAN outperformed KernelGAN.}
    \label{fig:results}
\end{figure}

\begin{figure}[htbp]
    \centering
    \includegraphics[width=.75\textwidth]{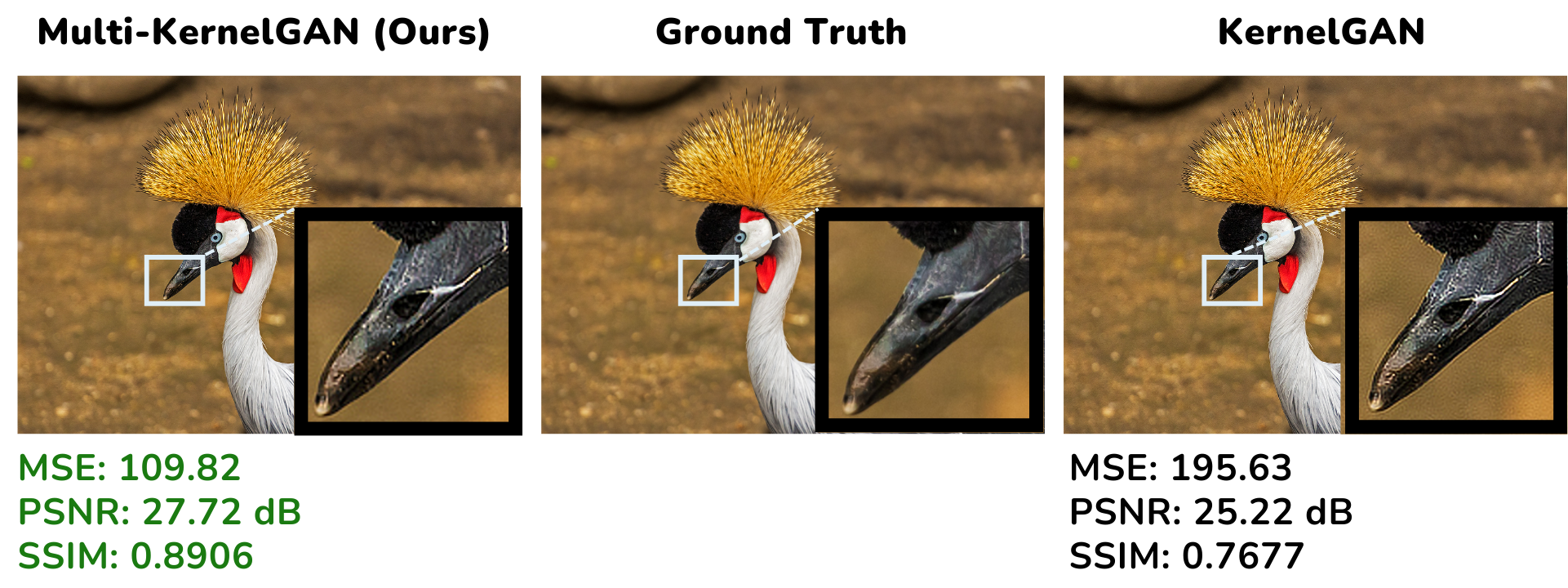}
    \caption{Illustration of the advantage of Multi-KernelGAN for image super-resolution as can be seen in the highlighted areas.}
    \label{fig:good_res}
\end{figure}

\begin{table}[ht]
    \centering
    \caption{Average PSNR, SSIM, and MSE for Multi-KernelGAN+ZSSR and KernelGAN+ZSSR.}
    \begin{tabular}{|l|c|c|c|}
        \hline
        \rowcolor{green!30}
        \textbf{Method}            & \textbf{Average PSNR} & \textbf{Average SSIM} & \textbf{Average MSE} \\ \hline
        Multi-KernelGAN+ZSSR       & 26.4559               & 0.8214                & 180.5488             \\ \hline
        KernelGAN+ZSSR             & 26.1906               & 0.8013                & 228.1127             \\ \hline
    \end{tabular}
    \label{tab:results}
\end{table}

\subsection{Performance Comparison}
The results in Table \ref{tab:results} show that \textit{Multi-KernelGAN+ZSSR} outperformed \textit{KernelGAN+ZSSR} across all metrics. Specifically, \textit{Multi-KernelGAN+ZSSR} achieved a higher average PSNR of 26.4559 dB compared to 26.1906 dB for \textit{KernelGAN+ZSSR}. Similarly, the average SSIM for \textit{Multi-KernelGAN+ZSSR} was 0.8214, surpassing the 0.8013 observed for \textit{KernelGAN+ZSSR}. Furthermore, \textit{Multi-KernelGAN+ZSSR} demonstrated a lower average MSE, indicating better reconstruction quality.

\section{Conclusion}
\subsection{Object Texture Constraints}
\begin{figure}[htbp]
    \centering
    \includegraphics[width=.75\textwidth]{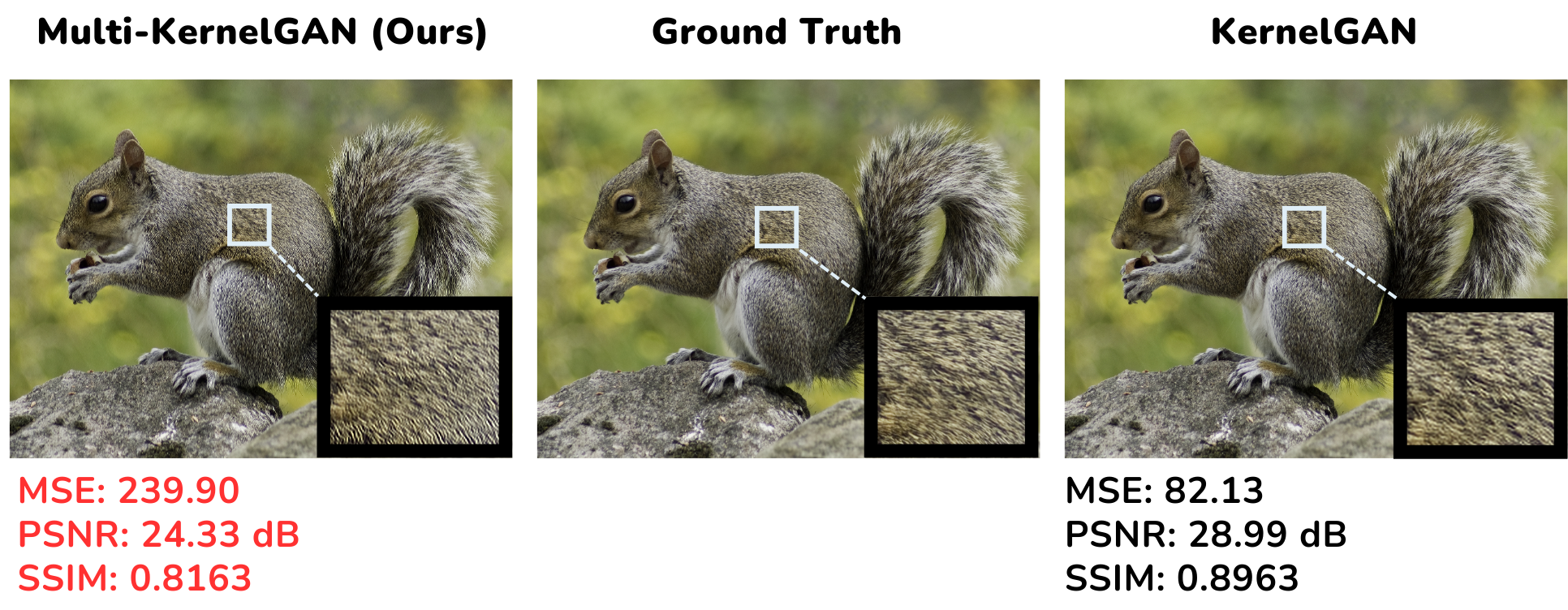}
    \caption{Illustration of the difficulty of Multi-KernelGAN with handling highly textured areas.}
    \label{fig:bad_res}
\end{figure}

Highly textured or noisy objects, such as animals with fur or those exhibiting rapid color changes, tend to cause KernelGAN to generate unstable or inaccurate kernels. Our observations indicate that these characteristics frequently result in bizarre and inconsistent kernel estimations across different iterations. In the case of multi-kernel GANs, this instability is exacerbated, as the presence of multiple kernels can make the object even more unstable, as can be seen in \autoref{fig:bad_res}. This can lead to poorer performance compared to the original KernelGAN, where the background effectively helps refine a single kernel. Consequently, the overall quality of the generated output may deteriorate, resulting in worse estimations than those produced by the simpler KernelGAN framework.

\subsection{Segmentation Size Limitation}
A significant limitation of our approach is the requirement for a minimum size for each segmented region. For KernelGAN and ZSSR to learn effectively, each segment must be sufficiently large to provide a meaningful "area to learn from." When segments are too small, the foundational principles of these algorithms become ineffective.

\subsection{Ideal Use Cases}
Our approach works best when applied to images containing distinct objects with clear boundaries and backgrounds. However, in more complex cases where objects are not clearly defined, or when their textures are too noisy (as discussed earlier), segmentation becomes ineffective. These factors significantly limit the broader applicability of multi-kernel estimation to certain types of images.

\subsection{Dataset Limitation}
A key limitation of our study is that the dataset was generated using a single, random blur kernel. As a result, our approach is tailored to perform best with that particular kernel. While this specificity may enhance performance in controlled scenarios, it presents challenges in real-world applications where reference data for evaluation is often unavailable. In such cases, traditional metrics like \textit{Blind/Referenceless Image Spatial Quality Evaluator} (BRISQUE) \cite{BRISQUE} and \textit{Perception based Image Quality Evaluator} (PIQUE) \cite{PIQUE} may be employed; however, our evaluation found that these metrics did not correlate with reference comparisons in all cases.

\section{Future Work}
\subsection{KernelGAN Instability}
One of the primary challenges faced by KernelGAN is instability during the training process. This instability can lead to inconsistent performance and unreliable kernel estimations. In our experiments, we found that achieving reliable results often required running the training process several times, which is not efficient and increases computational overhead. To address these issues, various techniques have been proposed, such as E-KernelGAN \cite{E-KernelGAN} and TVG-KernelGAN \cite{TVG-KernelGAN}, which aim to enhance the robustness of kernel estimation. Future work should not only implement these existing methods but also explore novel strategies that may further stabilize the training process. This could include optimizing hyperparameters, employing different loss functions, or integrating regularization techniques that help mitigate oscillations in the learning process.

\subsection{More Than Two Kernels Estimation}
Our current binary division approach relies on segmentation to estimate two distinct kernels. While this method has shown promise, there is significant potential to expand this approach to accommodate the estimation of multiple kernels. Future research can investigate advanced segmentation techniques that enable the handling of more than two kernels simultaneously, allowing for a more comprehensive analysis of complex scenes. This could involve developing algorithms that can identify and learn from varying textures and features within an image, leading to more nuanced and accurate kernel estimations.

\subsection{Stronger Discriminator}
The performance of KernelGAN can be significantly influenced by the quality of its discriminator. Future work should focus on enhancing the discriminator's architecture by incorporating additional layers to improve its capacity to learn complex patterns. Techniques such as attention mechanisms could be particularly beneficial, as they allow the model to focus on relevant parts of the input image, leading to more precise evaluations of the generated kernels.


\begin{thebibliography}{1}

    \bibitem{KernelGAN}
        \author{Sefi Bell-Kligler, Assaf Shocher and Michal Irani.}
        \date{7 Jan 2020.}
        \textit{Blind Super-Resolution Kernel Estimation using an Internal-GAN}:
        \url{https://arxiv.org/abs/1909.06581}

    \bibitem{YOLO}
        \author{Glenn Jocher and Jing Qiu.}
        \date{2024.}
        \textit{Ultralytics YOLOv8}:
        \url{https://github.com/ultralytics/ultralytics}

    \bibitem{SAM}
        \author{Meta AI.}
        \date{5 Apr 2023.}
        \textit{Segment Anything Model}:
        \url{https://arxiv.org/abs/2304.02643}

    \bibitem{SISR}
        \author{Daniel Glasner, Shai Bagon Michal and Irani.}
        \date{Nov 2009.}
        \textit{Super-Resolution from a Single Image}:
        \url{https://www.wisdom.weizmann.ac.il/~vision/single_image_SR/files/single_image_SR.pdf}

    \bibitem{ZSSR}
        \author{Assaf Shocher, Nadav Cohen and Michal Irani.}
        \date{17 Dec 2017.}
        \textit{"Zero-Shot" Super-Resolution using Deep Internal Learning}:
        \url{https://arxiv.org/abs/1712.06087}

    \bibitem{GANs}
        \author{Ian J. Goodfellow, Jean Pouget-Abadie, Mehdi Mirza, Bing Xu, David Warde-Farley, Sherjil Ozair, Aaron Courville and Yoshua Bengio.}
        \date{10 Jun 2014.}
        \textit{Generative Adversarial Networks}:
        \url{https://arxiv.org/abs/1406.2661}

    \bibitem{RCNN}
        \author{Ross Girshick, Jeff Donahue, Trevor Darrell and Jitendra Malik.}
        \date{22 Oct 2014.}
        \textit{Rich feature hierarchies for accurate object detection and semantic segmentation}:
        \url{https://arxiv.org/pdf/1311.2524}

    \bibitem{FASTER_RCNN}
        \author{Shaoqing Ren, Kaiming He, Ross Girshick and Jian Sun.}
        \date{6 Jan 2016.}
        \textit{Faster R-CNN: Towards Real-Time Object Detection with Region Proposal Networks}:
        \url{https://arxiv.org/pdf/1506.01497}

    \bibitem{Detectron2}
        \author{Yuxin Wu, Alexander Kirillov, Francisco Massa, Wan-Yen Lo and Ross Girshick.}
        \date{Oct 2019.}
        \textit{Detectron2}:
        \url{https://github.com/facebookresearch/detectron2}

    \bibitem{DIV2K}
        \author{Eirikur Agustsson, and Radu Timofte.}
        \date{24 Aug 2017.}
        \textit{NTIRE 2017 Challenge on Single Image Super-Resolution: Dataset and Study}:
        \url{https://ieeexplore.ieee.org/document/8014884}

    \bibitem{BRISQUE}
        \author{Anish Mittal, Anush Krishna Moorthy, and Alan Conrad Bovik}
        \date{Dec 2012.}
        \textit{No-Reference Image Quality Assessment in the Spatial Domain}:
        \url{https://live.ece.utexas.edu/publications/2012/TIP%20BRISQUE.pdf}

    \bibitem{PIQUE}
        \author{N. Venkatanath, D. Praneeth, Bh. M. Chandrasekhar, S. S. Channappayya and S. S. Medasani.}
        \date{16 Apr 2015.}
        \textit{Blind Image Quality Evaluation Using Perception Based Features}:
        \url{https://ieeexplore.ieee.org/document/7084843}

    \bibitem{E-KernelGAN}
        \author{Youngsoo Kim, Jeonghyo Ha, Yooshin Cho and Junmo Kim}
        \date{25 Apr 2022.}
        \textit{Unsupervised Blur Kernel Estimation and Correction for Blind Super-Resolution}:
        \url{https://ieeexplore.ieee.org/document/9762718}

    \bibitem{TVG-KernelGAN}
        \author{Jongeun Park,Hansol Kim and Moon Gi Kang.}
        \date{4 Apr 2023.}
        \textit{Kernel Estimation Using Total Variation Guided GAN for Image Super-Resolution}:
        \url{https://doi.org/10.3390/s23073734}

    \bibitem{VDSR}
        \author{Jiwon Kim, Jung Kwon Lee and Kyoung Mu Lee.}
        \date{11 Nov 2016.}
        \textit{Accurate Image Super-Resolution Using Very Deep Convolutional Networks}:
        \url{https://arxiv.org/pdf/1511.04587}

    \bibitem{EDSR}
        \author{Bee Lim, Sanghyun Son, Heewon Kim, Seungjun Nah and Kyoung Mu Lee.}
        \date{10 Jul 2017.}
        \textit{Enhanced Deep Residual Networks for Single Image Super-Resolution}:
        \url{https://arxiv.org/pdf/1707.02921}


\end{thebibliography}
\end{document}